\documentclass{article}
\usepackage{spconf,amsmath,graphicx}

\usepackage{enumitem}
\usepackage[hyphens]{url}
\usepackage{tabularx}
\usepackage{booktabs}
\usepackage{float}
\usepackage{amssymb}
\setlist{nosep, leftmargin=14pt}

\usepackage{mwe} 

\newcommand\blfootnote[1]{%
  \begingroup
  \renewcommand\thefootnote{}\footnote{#1}%
  \addtocounter{footnote}{-1}%
  \endgroup
}

\title{Pneumonia detection on chest X-ray using radiomic features and contrastive learning}
\name{Yan Han$^{1}$ \qquad Chongyan Chen$^{2}$  \qquad Ahmed Tewfik$^{1}$
\qquad Ying Ding$^{2,\star}$ \qquad Yifan Peng$^{3,\star}$}

\address{$^{1}$ Cockrell School of Engineering, The University of Texas at Austin\\
    $^{2}$School of Information, The University of Texas at Austin\\
    $^{3}$Department of Population Health Sciences, Weill Cornell Medicine}
\begin{document}
%
\maketitle
\begin{abstract}
Chest X-ray becomes one of the most common medical diagnoses due to its noninvasiveness. The number of chest X-ray images has skyrocketed, but reading chest X-rays still has been manually performed by radiologists, which creates huge burnouts and delays. Traditionally, radiomics, as a subfield of radiology that can  extract a large number of quantitative features from medical images, demonstrates its potential to facilitate medical imaging diagnosis before the deep learning era. With the rise of deep learning, the explainability of deep neural networks on chest X-ray diagnosis remains opaque. In this study, we proposed a novel framework that leverages radiomics features and contrastive learning to detect pneumonia in chest X-ray. Experiments on the RSNA Pneumonia Detection Challenge dataset show that our model achieves superior results to several state-of-the-art models ($> 10\%$ in F1-score) and increases the model's interpretability.
\end{abstract}
\begin{keywords}
radiomics, medical imaging, CNN, chest X-ray, neural networks, interpretability
\blfootnote{$^*$ Co-corresponding.}
\end{keywords}
\section{Introduction}
\label{sec:intro}
Pneumonia is the leading cause of people hospitalized in the US \cite{jain2015community}. It requires timely and accurate diagnosis for immediate treatment. As one of the most ubiquitous diagnostic imaging tests in medical practice, chest X-ray plays a crucial role in pneumonia diagnosis in clinical care and epidemiological studies \cite{tang2020automated}. However, rapid pneumonia detection in chest X-rays is not always available, particularly in the low-resource settings where there are not enough trained radiologists to interpret chest X-rays. There is, therefore, a critical need to develop an automated, fast, and reliable method to detect pneumonia on chest X-rays.

With the great success of deep learning in various fields, deep neural networks (DNNs) have proven to be powerful tools that can detect pneumonia to  augment  radiologists \cite{wang2017chestx,jaiswal2019identifying,wang2020covid,kermany2018identifying}. However,  most of the DNNs lacks explainability due to their black-box nature. Thus researchers still have a limited understanding of DNNs’ decision-making process.

One method of increasing the explainability of DNNs in chest radiographs is to leverage radiomics. Radiomics is a novel feature transformation method for detecting clinically relevant features from radiological imaging data that are difficult for the human eye to perceive. It has proven to be a highly explainable and robust technique because it is related to a specific region of interest (ROI) of the chest X-rays \cite{chen2017development}. However, directly combining radiomic features and medical image hidden features provides only marginal benefits, a result mostly due to the lack of correlations at a ``mid-level''; it can be challenging to relate raw pixels to radiomic features. 
In efforts to make more efficient use of multimodal data, several recent studies have shown
promising results from contrastive representation learning \cite{zhang2020contrastive,chen2020simple}. But, to the best of our knowledge, no studies have exploited the naturally occurring pairing of images and radiomic data. 

In this study, we proposed a framework that leverages radiomic features and contrastive learning to detect pneumonia in chest X-ray. Our framework improves chest x-ray representations by maximizing the agreement between true image-radiomics pairs versus random pairs via a bidirectional contrastive objective between the image and human-crafted radiomic features. Experiments on the RSNA Pneumonia Detection Challenge dataset \cite{shih2019augmenting} show that our methods can fully utilize unlabeled data, provide a more accurate pneumonia diagnosis, and remedy the black-box’s transparency. 

Our contribution in this work is three-fold: (1) We introduce a framework for pneumonia detection that combines the expert radiographic knowledge (radiomic features) with deep learning. (2) We improve chest X-ray representations by exploring the use of contrastive learning. Our model thus has the advantages of utilizing the paired radiomic features requiring no additional radiologist input. (3) We find that our models significantly outperform baselines in pneumonia detection with improved model explainability.

\section{Related Work}
\label{sec:related}

Pneumonia detection is a binary classification task which requires to classify a chest radiology image into either pneumonia or normal. Popular pneumonia detection dataset includes RSNA Pneumonia Detection Challenge \cite{shih2019augmenting} and  pediatric pneumonia diagnosis\footnote{\url{https://data.mendeley.com/datasets/rscbjbr9sj/3}}. 

Traditionally, non-image features (e.g., patient age, gender, and body temperature) and radiomic features \cite{gillies2016radiomics} are used for automatic chest disease classification. In recent years, many studies explored deep neural networks (DNNs) for this task \cite{khan2020intelligent,rajpurkar2017chexnet,liang2020transfer}. For instance, Rajpurkar et al. introduced the CheXNet, a deep CNN trained to predict 14 diseases on chest X-ray \cite{rajpurkar2017chexnet}. Liang and Zheng used the Residual Neural Network (ResNet-18) \cite{he2016deep}	pre-trained on the NIH ChestX-ray 14 dataset and fine-tuned on the child’s chest X-rays dataset for pediatric pneumonia diagnosis \cite{liang2020transfer}.

For the medical image classification task, semi-supervised or unsupervised learning methods have benefited this task hugely because preparing annotated corpora is generally time-consuming and expensive. It also requires domain expertise and significant effort to ensure accuracy and consistency. To relieve this problem, one method is to utilize unlabeled image data. For example, Tang et al., \cite{tang2019tuna} introduces the task-oriented unsupervised adversarial network, which consists of a cyclic I2I translation framework for RSNA Pneumonia Detection Challenge and a pediatric pneumonia diagnosis dataset.

Another popular trend, especially in recent years, is the contrastive representation learning \cite{henaff2019data,he2020momentum,chen2020simple}. Nevertheless, it may not be beneficial to directly apply these visual contrastive learning methods to medical images than pre-training models on ImageNet and fine-tuning them on medical images, mainly because the medical images have high inter-class similarity \cite{zhang2020contrastive}. Thus, Zhang et al. \cite{zhang2020contrastive} proposed to use contrastive learning to learn visual representations from radiology images and text reports by maximizing the agreement between image-text representation pairs. Different from these works, we studied the contrastive learning between radiomics and convolutional neural networks (CNN) features to obtain medical visual representations. Therefore, our model does not require radiology text reports which are usually not publicly available. To this end, we deem that our framework is simple yet scalable when coupled with large-scale medical image datasets.

\section{Proposed Method}
\label{sec:method}

\begin{figure*}[h]
  \centering
  \includegraphics[width=.75\linewidth]{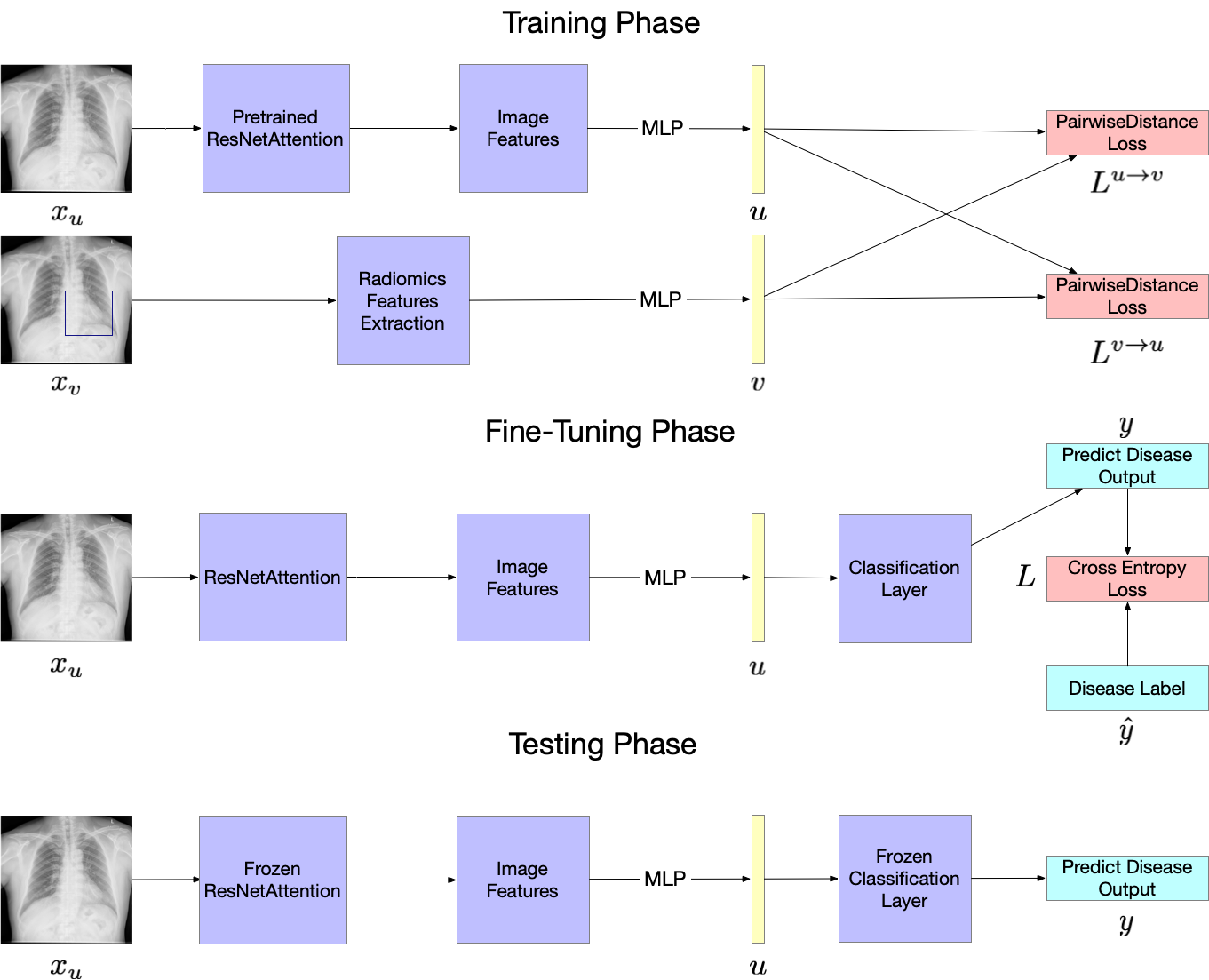}
  \caption{An overview of the proposed model.}
  \label{fig1}
\end{figure*}

Inspired by recent contrastive learning algorithms \cite{zhang2020contrastive}, our model learns representations by maximizing agreement between radiomics features related to pneumonia ROI of the chest X-rays and the image features extracted by the attention-based convolutional neural network (CNN) model, via a contrastive loss in the latent space. Since radiomics can be considered as the quantified prior knowledge of radiologists, we deem that our model is more interpretable than others. As illustrated in Figure \ref{fig1}, our framework consists of three phases: contrastive training, supervised fine-tuning, and testing.

\textbf{Contrastive training}. The model is given two inputs, $x_u$ and $x_v$. $x_u$ is the original chest X-rays without a corresponding paired bounding box. $x_v$ is the original chest X-rays with an additional paired bounding box. For normal chest X-rays, we take the whole image as a  bounding box. 

For $x_u$, we utilize the pre-trained attention-based CNN models, Residual Attention Network (ResNet-18Attention)  \cite{wang2017residual} pre-trained on CIFAR-10 \cite{krizhevsky2009learning}, as the backbone of the network. We replace the last fully-connected layer with a multilayer perceptron (MLP) to generate a 128-dimensional image features vector $u$.
For $x_v$, we apply the PyRadiomics\footnote{\url{https://pyradiomics.readthedocs.io/en/latest/}} to extract 102-dimensional quantitative features, and \cite{van2017computational} showed the details of these quantitative features and extraction process. We then use an MLP to map the features to a 128-dimensional radiomics feature $v$.

At each epoch of training, we sample a mini-batch of $N$ input pairs $(X_u, X_v)$ from the training data, and calculate their image features and radiomics features pairs $(U, V)$. We use $(u_i, v_i)$ to denote the $i$th pair. The training loss function will be divided into two parts. The first part is a contrastive image-to-radiomics loss:
\begin{equation}
    L^{u\rightarrow v}_i = -\log \frac{\exp(<u_i, v_i>/\tau)}{\sum_{k=1}^{N}\exp(<u_i, v_k>/\tau)}
\end{equation}
where $<u_i, v_i>$ represents the pairwise distance, i.e. $[\sum(u_i - v_i)^p]^{\frac{1}{p}}$ and $p$ represents the norm degree, e.g., $p=1$ and $p=2$ represent the Taxicab norm and Euclidean norm, respectively; and $\tau \in \mathbb{R}^+$ represents a temperature parameter. In our model, we set $p$ to 2 and $\tau$ to 0.1. Like previous work \cite{zhang2020contrastive}, which uses a contrastive loss between inputs of the different modalities, our image-to-radiomics contrastive loss is also asymmetric for each input modality. We thus define a similar radiomics-to-image contrastive loss as:
\begin{equation}
    L^{v\rightarrow u}_i = -\log \frac{\exp(<v_i, u_i>/\tau)}{\sum_{k=1}^{N}\exp(<v_i, u_k>/\tau)}
\end{equation}
Our final loss is then computed as a weighted combination of the two losses averaged over all pairs in each minibatch where $\lambda \in [0, 1]$ is a scalar weight
\begin{equation}
    L_{train} = \frac{1}{N} \sum_{i=1}^N (\lambda L^{u\rightarrow v}_i + (1 - \lambda)L^{v\rightarrow u}_i)
\end{equation}

\textbf{Supervised fine-tuning}. We follow the work of Zhang et al. \cite{zhang2020contrastive} by fine-tuning both the CNN weights and the MLP blocks together, which closely resembles how the pre-trained CNN weights are used in practical applications. In this process, the loss function is the cross-entropy loss where $\hat{y}$ and $y$ represent the true and predicted disease label, respectively.:
\begin{equation}
L_{fine-tune} = -(\hat{y}\log y + (1-\hat{y})\log(1-y))
\end{equation}

\textbf{Testing}. The model is only given one input, the original chest X-rays $x_u$ without a corresponding paired bounding box. Image features are extracted then mapped into the 128-dimensional feature representation $u$. Finally, the predicted output is calculated based on $u$.

\section{Experiments and discussion}
\label{sec:result}

\subsection{Dataset and Experimental Settings}
To evaluate the performance of our proposed  model, we conducted experiments on a public Kaggle dataset: RSNA Pneumonia Detection Challenge\footnote{\url{https://www.kaggle.com/c/rsna-pneumonia- detection-challenge/data}}. It contains 30,227 frontal-view images, out of which 9,783 images has pneumonia with a corresponding bounding box. We used 75\% imaged for training and fine-tuning and 25\% for testing.

We used SGD as our optimizer and set the initial learning rate as 0.1. We iterated the training and fine-tuning process for 200 epochs with batch size 64 and early stooped if the loss did not decrease.
We reported accuracy, F1 score, and the area under the receiver operating characteristic curve (AUC).


\subsection{Results}

We compared four models: (1) ResNet-18, (2) ResNet-18 with radiomics features (ResNet-18Radi), (3) ResNet-18 with the attention mechanism (ResNet-18Att), and (4) ResNet-18Attention with radiomics features (ResNet-18AttRadi).

Experimental results are shown in Table \ref{tab1}. Compared with the baseline models (ResNet-18 and ResNet-18Att), our radiomics-based models (ResNet-18Radi and ResNet-18AttRadi) achieved better performance on the pneumonia/normal binary classification task. It suggests that radiomic features can provide additional strengths over the image features extracted by the CNN model.
Compared ResNet-18Att with ResNet-18 and ResNet-18AttRadi with ResNet-18Radi, we observed that the attention mechanism could effectively boost the classification accuracy. It proves our hypothesis that pneumonia is often related to some specific ROI of chest X-rays. Hence, the attention mechanism makes it easier for the CNN model to focus on those regions. 
\begin{table}[!htpb]
\vspace{-1em}
  \caption{Experimental results}
  \label{tab1}
  
\centering
\begin{tabularx}{\linewidth}{Xccc}

\toprule
\textbf{Model}            & \textbf{Accuracy} & \textbf{F1 score} & \textbf{AUC} \\ \midrule
ResNet-18                    & 0.763             & 0.782             & 0.795        \\
ResNet-18Att           & 0.815             & 0.826             & 0.848        \\
ResNet-18Radi          & 0.851                 & 0.901                 & 0.898            \\
ResNet-18AttRadi & \textbf{0.886}                 & \textbf{0.927}                 & \textbf{0.923}            \\
\bottomrule
\end{tabularx}
\end{table}

\begin{figure}[!htbp]
  \centering
  \includegraphics[width=.9\linewidth,clip,trim=0 0 0 3em]{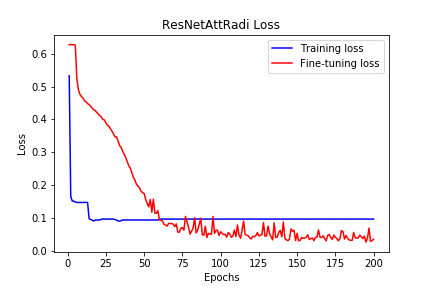}
  \vspace{-1.5em}
  \caption{The training and fine-tuning loss convergence for the ResNet-18AttRadi model.}
  \label{fig2}
\end{figure}

Figure \ref{fig2} shows the training and fine-tuning loss convergence for the ResNet-18AttRadi model on the training set. We find that the loss drops rapidly during the pre-training stage within just a few epochs, revealing that contrastive learning makes the model learn to extract image features fast and effectively. 

To fairly evaluate the impact of radiomics features on ROI, we conducted additional experiments using the whole image as a bounding box to extract the radiomics features, denoted as ResNet-18FairRadi and ResNet-18AttFairRadi. Table \ref{tab2} shows that even if without ROI, the radiomics features could improve the performance of the deep learning model by 5\% in F1 score. This observation further demonstrates that combining radiomics features with a deep learning model for reading chest X-rays is necessary. 
\begin{table}[!htbp]

  \caption{Experimental Results Without Using Bounding Box}
  \label{tab2}
  
\centering
\begin{tabularx}{\linewidth}{Xccc}

\toprule
\textbf{Model}            & \textbf{Accuracy} & \textbf{F1 score} & \textbf{AUC} \\ \midrule
ResNet-18                    & 0.763             & 0.782             & 0.795        \\
ResNet-18FairRadi          & 0.821                 & 0.841                 & 0.864            \\\midrule
ResNet-18Att           & 0.815             & 0.826             & 0.848        \\
ResNet-18AttFairRadi & \textbf{0.854}                 & \textbf{0.884}                 & \textbf{0.877}            \\ \bottomrule
\end{tabularx}
\end{table}

\subsection{Visualization of the deep learning model}
\begin{figure}
  \centering
  \includegraphics[width=\linewidth,clip,trim=0 0 0 3em]{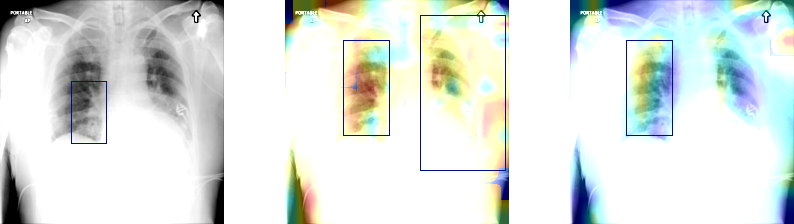}
  \vspace{-2em}
  \caption{An example of visualization of attention maps. The left figure is the original \textbf{Pneumonia} chest X-ray with a bounding box. The right two figures are the attention maps of the final attention layer ResNet-18Att and ResNet-18AttRadi, respectively.}
  \label{fig:res}
\end{figure}

To demonstrate the interpretability of our model, we show some selected examples of model visualization, i.e., attention maps of ResNet-18Att and ResNet-18AttRadi. Figure \ref{fig:res} shows the original chest X-ray with a bounding box, attention map of the final attention layer of the ResNet-18Att and ResNet-18AttRadi, respectively. These examples suggest that our ResNet-18AttRadi model can focus on a more accurate area of the chest X-ray while ResNet-18Att attends to almost the whole image and contains plenty of attention noise.  This illustrates that contrastive learning can help the model learn from radiomics features related to certain ROIs and thus attend more to the correct regions. And more examples of the attention maps can be found in the supplemental material. 

\section{Conclusion and future work}
\label{sec:conclusion}
In this work, we present a novel framework by combining radiomic features and contrastive learning to detect pneumonia from chest X-ray. Experimental results showed that our proposed models could achieve superior performance to baselines. We also observed that our model could benefit from the attention mechanism to highlight the ROI of chest X-rays.

There are two limitations to this work. First, we evaluated our framework on one deep learning model (ResNet). We plan to assess the effect of radiomic features on other DNNs in the future. Second, our model relies on bounding box annotations during the training phase. We plan to leverage weakly supervised learning to automatically generate bounding boxes on large-scale datasets to ease the expert annotating process. In addition, we will compare contrastive learning with multitask learning to further exploit the integration of radiomics with deep learning.

While our work only scratches the surface of contrastive learning using radiomics knowledge in the medical domain, we hope it will shed light on the development of explainable models that can efficiently use domain knowledge for medical image understanding.

\section{Potential Negative Societal Impact}
This research study was conducted retrospectively using human subject data made available \cite{shih2019augmenting}. We don't 

\section{Acknowledgments}
\label{sec:acknowledgments}
This project was supported by the National Library of Medicine under award number 4R00LM013001 and Amazon Machine Learning Grant. Ying Ding receives research support from Amazon. Yifan Peng is a coinventor on patents awarded and pending.




\bibliographystyle{IEEEbib}
\bibliography{strings,refs}

\setcounter{figure}{0}
\renewcommand{\figurename}{Supplementary Figure}
\renewcommand{\thefigure}{S\arabic{figure}}

\newpage
\begin{figure*}[!htbp]
  \centering
  \includegraphics[width=\linewidth,clip,trim=0 0 0 3em]{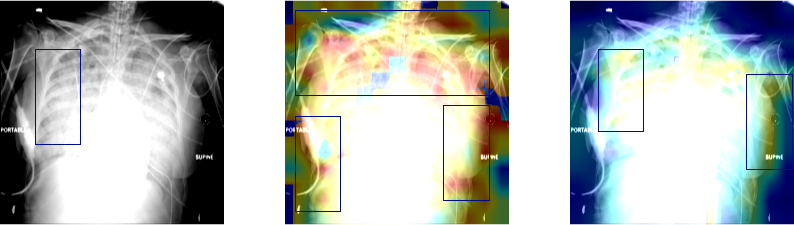}
  \vspace{-2em}
  \caption{Visualization of Attention Maps. The left figure is the original \textbf{Pneumonia} chest X-ray with a bounding box, and the middle figure and right figure are the attention maps of the final attention layer ResNet-18Att and ResNet-18AttRadi, respectively.}
\end{figure*}

\begin{figure*}[!htbp]
  \centering
  \includegraphics[width=\linewidth,clip,trim=0 0 0 3em]{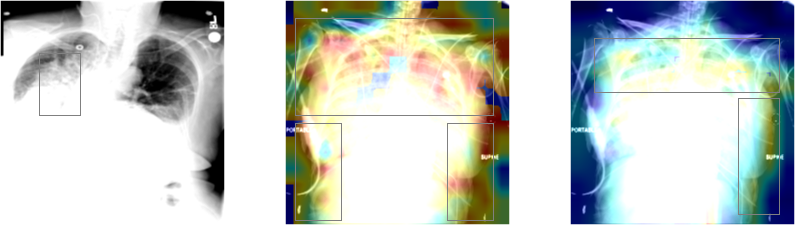}
  \vspace{-2em}
  \caption{Visualization of Attention Maps. The left figure is the original \textbf{Pneumonia} chest X-ray with a bounding box, and the middle figure and right figure are the attention maps of the final attention layer ResNet-18Att and ResNet-18AttRadi, respectively.}
\end{figure*}

\begin{figure*}[!htbp]
  \centering
  \includegraphics[width=\linewidth,clip,trim=0 0 0 3em]{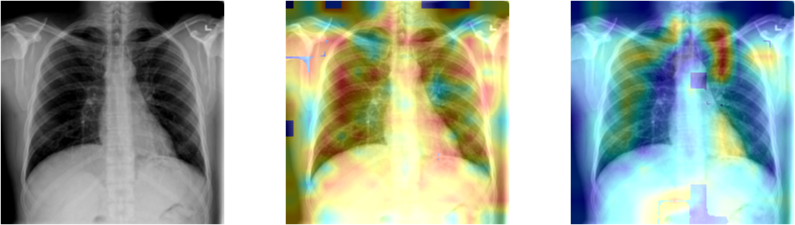}
  \vspace{-2em}
  \caption{Visualization of Attention Maps. The left figure is the original \textbf{Normal} chest X-ray, and the middle figure and right figure are the attention maps of the final attention layer ResNet-18Att and ResNet-18AttRadi, respectively.}
\end{figure*}

\begin{figure*}[!htbp]
  \centering
  \includegraphics[width=\linewidth,clip,trim=0 0 0 3em]{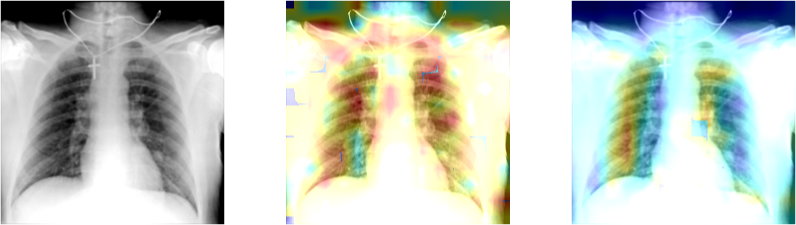}
  \vspace{-2em}
  \caption{Visualization of Attention Maps. The left figure is the original \textbf{Normal} chest X-ray, and the middle figure and right figure are the attention maps of the final attention layer ResNet-18Att and ResNet-18AttRadi, respectively.}
\end{figure*}

\end{document}